\documentclass[letterpaper, 10 pt, journal, twoside]{IEEEtran}
\usepackage{graphicx}
\usepackage{amsmath}
\usepackage{comment}
\usepackage{array}
\usepackage{booktabs}
\usepackage{amssymb}
\usepackage{pifont}
\usepackage{siunitx}
\usepackage{threeparttable}
\usepackage[caption=false,font=footnotesize]{subfig}
\usepackage{multirow}
\usepackage{cite}
\usepackage{url}
\newcommand{\cmark}{\ding{51}}
\newcommand{\xmark}{\ding{55}}
\usepackage{fancyhdr}

\newcommand{\RALEVENHEADER}{}
\newcommand{\RALODDHEADER}{JUSTDEPTH: REAL-TIME RADAR-CAMERA DEPTH ESTIMATION WITH SINGLE-SCAN LIDAR SUPERVISION}

\fancypagestyle{firstpage}{%
  \fancyhf{}
  
  \fancyhead[LO]{\footnotesize\RALEVENHEADER}
  \fancyhead[RO]{\footnotesize\thepage}
}

\begin{document}

\title{JustDepth: Real-Time Radar-Camera Depth Estimation with Single-Scan LiDAR Supervision}

\author{Wooyung~Yun$^{1}$, Dongwook~Kim$^{1}$, and Soomok Lee$^{2,\dagger}$%

\thanks{$^{1}$Wooyung Yun and Dongwook Kim are with the Department of Artificial Intelligence, Ajou University, Suwon 16499, Republic of Korea
(e-mail: \{woodolly17, jayk01213\}@ajou.ac.kr).
$^{2}$Soomok Lee is the Corresponding author$^\dagger$ and is the Faculty of Department of Computer Science, Kennesaw State University, Marietta GA, 30060, United States  
(e-mail: slee337@kennesaw.edu).}%
\thanks{Code: \protect\url{https://github.com/TPyun/JustDepth}.} 
}

\maketitle
\thispagestyle{firstpage}
\pagestyle{fancy}
\bstctlcite{BSTcontrol:no_dash}

\begin{abstract}
Accurate yet low-latency depth is essential for radar-camera perception in autonomous systems. Cameras provide rich appearance but lack metric scale, whereas automotive radar offers metric range but is sparse and noisy. Many pipelines are multi-stage or depend on auxiliary annotations, increasing latency and limiting portability. 
We introduce JustDepth, a single-stage radar-camera depth estimator trained only with radar, camera, and single-scan LiDAR. All radar returns are aggregated into a fixed-width 1D representation, decoupling runtime from point count. A Height Fusion Block fuses modalities, a lightweight GNN propagates depth globally, and a training-only confidence decoder stabilizes learning with zero test-time cost. We mitigate stripe artifacts via simple augmentations and quantify them using the Vertical-Horizontal Gradient Ratio (VHGR). On nuScenes, compared to recent state-of-the-art methods, JustDepth maintains accuracy while reducing inference time by $39.7\times$ and stripe artifacts by $66\%$ as measured by VHGR.
\end{abstract}


\section{Introduction}
\IEEEPARstart{A}{ccurate} 3D reconstruction is crucial for autonomous driving and robotics. LiDAR provides precise depth but degrades in adverse weather, while automotive mmWave radar is weather-robust yet sparse and noisy with limited angular resolution. Monocular images provide dense appearance cues but lack metric scale. Fusing radar with monocular images enables dense, metric-scale depth that remains robust in challenging conditions.

\begin{figure}[t]
 \centering
 \includegraphics[width=0.99\columnwidth,trim=0 0 0 0,clip]{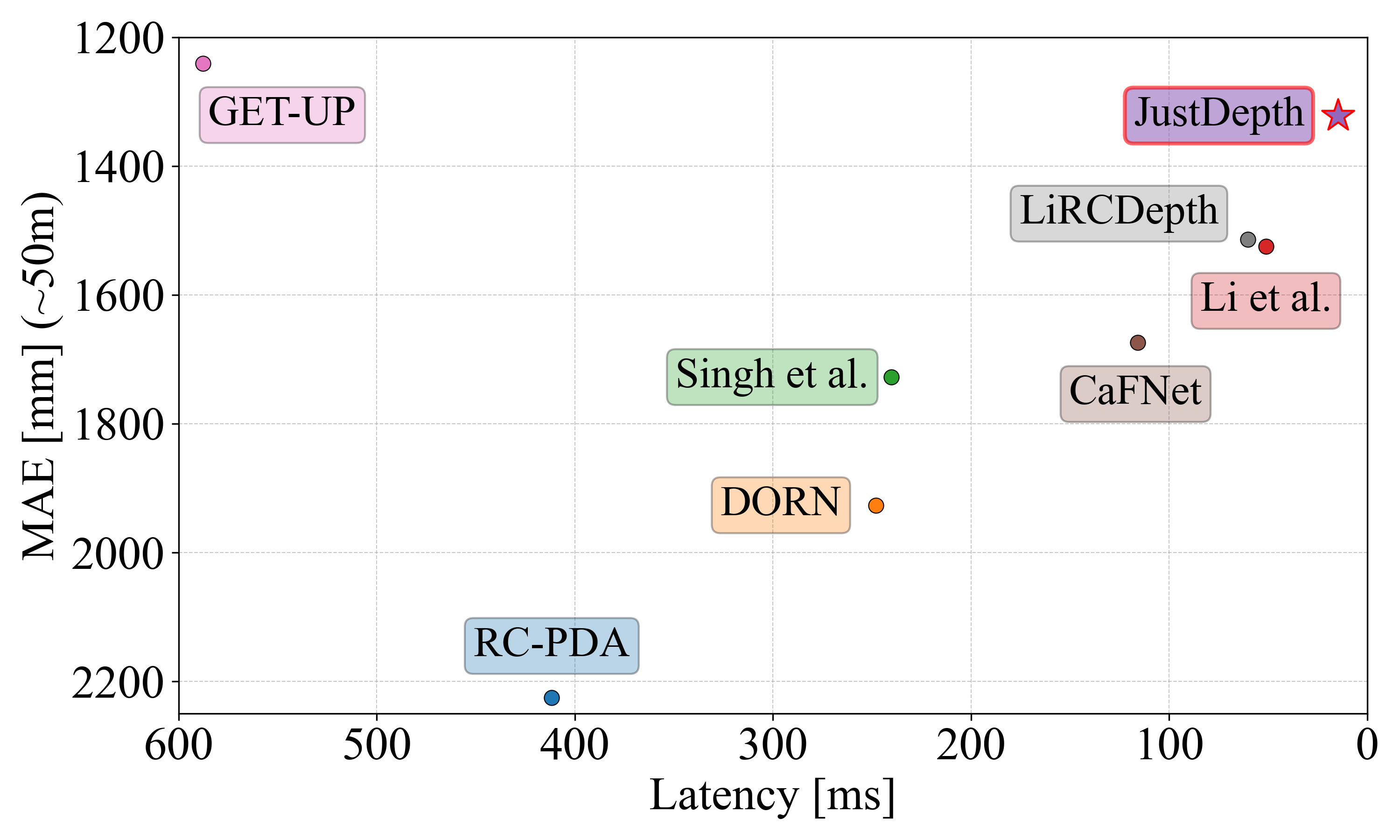}
 \vspace{-24pt}
\caption{\textbf{Error-latency trade-off of radar-camera depth estimation methods.}
 Each point shows MAE at 0-50\,m versus per-frame latency (ms) on the nuScenes test set~\cite{caesar2020nuscenes}, measured on an NVIDIA RTX~4070~Ti. JustDepth with an 8-layer GNN achieves real-time inference ($\approx$14.8\,ms) while maintaining accuracy comparable to prior methods.}
 \label{fig:performance_graph}
\end{figure}

\begin{figure*}[!t]
\vspace{5pt}
\centering
  \includegraphics[width=0.92\textwidth,trim=0 0 0 0,clip]{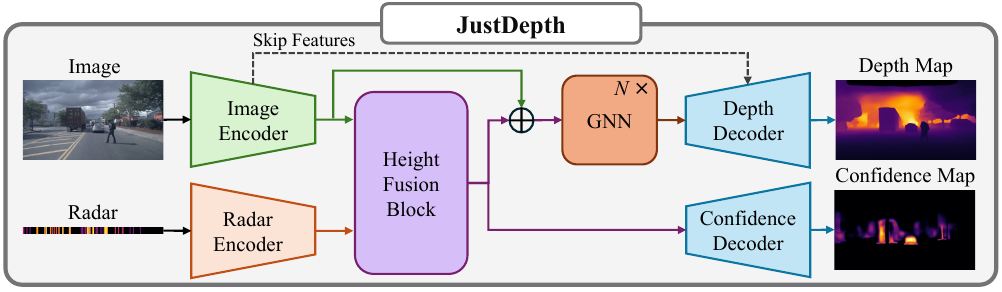}
\vspace{-10pt}
\caption{\textbf{Overview of the JustDepth architecture.} The network takes an RGB image and a 1D radar scan as inputs. An Image Encoder and a Radar Encoder extract features. These are fused by the Height Fusion Block and combined with the final image features before being processed by an $N$-layer GNN. The resulting features are decoded by the Depth Decoder to produce the final dense depth map, while the Confidence Decoder predicts a per-pixel reliability map during training (discarded at inference).}
\label{fig:arch}
\end{figure*}

Prior work in radar-camera depth estimation has become increasingly complex in both architecture and supervision. On the architectural side, many methods adopt multi-stage pipelines or add external modules~\cite{li2024radarcam,lin2020depth,DORN_radar,RCDPT,long2021radar,singh2023depth,sun2024cafnet}, often first constructing an intermediate sparse or quasi-dense depth map and then refining it with a second network. Some approaches also integrate a pre-trained monocular depth model~\cite{li2024radarcam}, which increases latency and couples performance to external pretraining. In terms of supervision and annotation cost, many approaches rely on auxiliary labels such as semantic or panoptic masks and 2D/3D bounding boxes~\cite{li2024semantic,li2024sparse,long2021radar,gasperini2021r4dyn,singh2023depth,li2024radarcam,sun2024cafnet,sun2025lircdepth,sun2025getup} to construct LiDAR ground truth by accumulating multiple scans or to complement single-scan LiDAR supervision, which raises annotation cost and hinders portability across datasets and sensors.

In this study, we introduce JustDepth, a single-stage radar-camera depth estimator. The network takes an RGB image and a radar scan as input and predicts dense depth directly, without intermediate products~\cite{li2024radarcam,lin2020depth,DORN_radar,RCDPT,long2021radar,singh2023depth,sun2024cafnet} or a pretrained monocular module~\cite{li2024radarcam}. All radar returns in a frame are compressed into a fixed-width 1D representation, which keeps computation and latency effectively constant with respect to the number of points. A Height Fusion Block aligns radar and image features. From the fused feature, two decoders operate in parallel: a graph neural network followed by a depth decoder that propagates depth cues across the scene to produce a dense map, and a training-only confidence decoder that learns per-pixel reliability and is discarded at inference.

Unlike pipelines that depend on auxiliary segmentation labels or bounding boxes to accumulate multi-sweep LiDAR or provide semantic supervision~\cite{long2021radar,singh2023depth,li2024radarcam,sun2024cafnet,sun2025getup,sun2025lircdepth,li2024sparse}, JustDepth trains with a single LiDAR sweep and no auxiliary annotations. However, single-scan supervision is still prone to LiDAR Distribution Leakage~\cite{li2024sparse}, where predictions inherit scanline stripes. To mitigate this, we use point upsampling and a synchronized rotation augmentation that applies the same random angle to the image, radar, and LiDAR. To quantify the effectiveness of this mitigation, we evaluate the Vertical-Horizontal Gradient Ratio (VHGR), which measures the imbalance between vertical and horizontal gradients. When evaluated on nuScenes~\cite{caesar2020nuscenes} with raw single-scan LiDAR as sparse ground truth and no auxiliary annotations, JustDepth attains competitive accuracy and the fastest per-frame latency among the compared methods. The resulting latency-accuracy trade-off is summarized in Fig.~\ref{fig:performance_graph}.

Our main contributions are summarized as:
\begin{itemize}
\item We propose a single-stage, constant-latency architecture that encodes all radar returns into a fixed-width 1D representation, fuses them with image features via height-wise self-attention, and uses a GNN for global propagation.
\item We introduce a training-only confidence decoder that directly supervises the fusion module to localize radar-supported pixels without producing an explicit intermediate depth map, and is discarded at inference, improving accuracy without adding stages or latency at test time.
\item We propose an LDL mitigation strategy that uses point upsampling and rotation with reflection padding applied simultaneously to the image, radar, and LiDAR, and we introduce VHGR to quantify stripe artifacts.
\end{itemize}


\begin{figure*}[t]
\vspace{5pt}
\centering
  \includegraphics[width=0.92\textwidth,trim=0 0 0 0,clip]{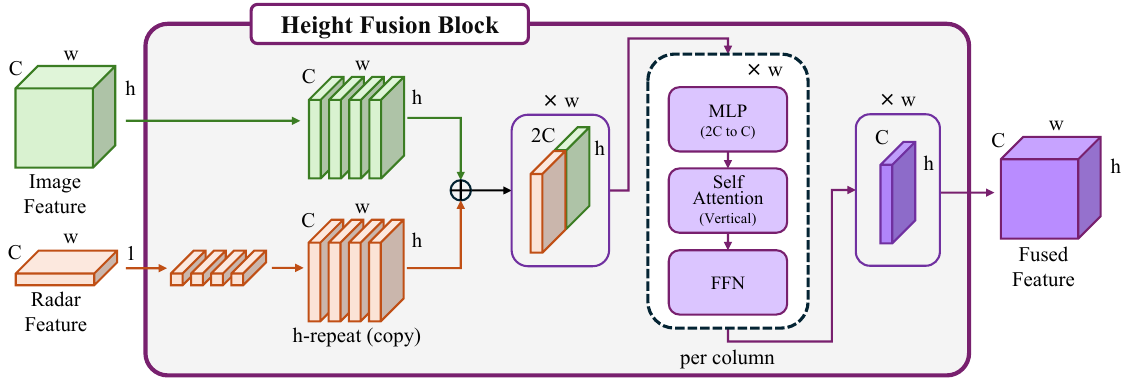}
\vspace{-10pt}
\caption{\textbf{Height Fusion Block.} The block takes image features $i_L\in\mathbb{R}^{B\times C\times h\times w}$ and radar features $r_L\in\mathbb{R}^{B\times C\times w}$, repeats $r_L$ along the height dimension to form $h\times C$ tokens, and concatenates them with the corresponding column to produce $2C$-dimensional per-column tokens. Each column is then processed by a per-column linear projection, followed by height-wise self-attention and a feed-forward network. Finally, the refined tokens are reshaped back to $\mathbb{R}^{B\times C\times h\times w}$ and passed through a convolutional module to produce the fused feature.}
\label{fig:hfb}
\end{figure*}

\section{Related Work}

\noindent\textbf{Monocular Depth Estimation.}
Recent monocular depth methods increasingly adopt transformer backbones and large-scale training to produce more coherent predictions, e.g., DPT~\cite{ranftl2021vision} and Depth Anything~\cite{yang2024depth}. 
Metric depth and camera geometry are addressed by explicitly modeling camera space and intrinsics in works such as Metric3D~\cite{yin2023metric3d} and UniDepth~\cite{piccinelli2024unidepth}. 
Diffusion priors have also been explored to refine structure and details, e.g., Marigold~\cite{ke2024repurposing}. 
DepthPro presents a foundation model for zero-shot metric monocular depth, producing high-resolution predictions with an efficient multi-scale transformer~\cite{bochkovskii2025depthpro}.

\noindent\textbf{Radar-Camera Depth Estimation.}
For automotive depth estimation, many radar-camera approaches first build an intermediate sparse or quasi-dense depth map from radar and images, then refine it with a second network~\cite{lin2020depth,DORN_radar,long2021radar,singh2023depth,sun2024cafnet,li2024radarcam}. Such multi-stage or per-point/per-ROI pipelines often incur high latency. Recent methods also plug a pre-trained monocular depth model into the pipeline~\cite{li2024radarcam,wang2025tacodepth}. They use monocular depth either as an input or as additional supervision and achieve their best results in this setting, which further increases system complexity and data dependence.

To cope with sparse and noisy measurements, several methods exploit auxiliary labels and multi-frame LiDAR accumulation. Semantic or panoptic masks and 2D/3D bounding boxes are used to filter radar points and to construct dense LiDAR ground truth~\cite{li2024semantic,li2024sparse,gasperini2021r4dyn,long2021radar,singh2023depth,sun2024cafnet,sun2025getup,sun2025lircdepth}. These strategies rely on accurate ego pose, temporally contiguous sequences, and high-quality annotations. Li \emph{et al.}~\cite{li2024sparse} reduce this burden by supervising on a single LiDAR sweep but still depend on image semantic segmentation. In contrast, JustDepth uses single-scan LiDAR without auxiliary labels or monocular depth and focuses on efficient single-stage fusion.

\section{Method}
JustDepth fuses image and radar features for depth estimation through a novel height-based fusion block and propagates globally via graph neural network layers, as illustrated in Fig.~\ref{fig:arch}. This architecture comprises a 2D image encoder and a 1D radar encoder that compresses each scan into a fixed-width projection, making the radar branch latency constant with respect to the number of radar returns. Their features are fused in a Height Fusion Block, passed through a GNN that propagates depth cues globally, and finally decoded to depth and confidence. We train on single-scan LiDAR and use augmentations to reduce LiDAR Distribution Leakage.

\subsection{Fusion Method}
The fusion stage aligns and merges image features and the projected radar feature into a unified representation, as illustrated in Fig.~\ref{fig:hfb}. We first introduce the final latents from each encoder. \(H\) and \(W\) denote the height and width of the input image and radar projection, while \(h\) and \(w\) denote the height and width of the latent feature map after downsampling.

\noindent\textbf{Image Encoder.} 
A ResNet-style backbone~\cite{he2016deep} processes the RGB input into a hierarchy of feature maps. The output of its final stage is denoted 
\(
i_L \in \mathbb{R}^{B \times C \times h \times w},
\)
where \(B\) is batch size and \(C\) the channel dimension.

\noindent\textbf{Radar Encoder.}
We project calibrated radar points onto the image plane and retain only the horizontal pixel coordinate \(u \in \{0,\dots,W{-}1\}\) and the range, discarding the vertical coordinate and other attributes (e.g., Doppler, RCS). We then rasterize along the horizontal axis into a 1D grid of width \(W\). For each column \(u\), we store the minimum range of all points falling into that column and set empty columns to zero.
This produces a fixed-width 1D radar scan of shape \(B \times 1 \times W\) without any vertical interpolation. A one-dimensional convolutional backbone processes this scan and downsamples it to width \(w\), producing the latent radar feature \(r_L \in \mathbb{R}^{B \times C \times w}\); since \(W\) and \(w\) are fixed, the encoder’s computation is independent of the number of raw radar points.

\noindent\textbf{Height Fusion Block.}
The Height Fusion Block takes the final image features \(i_L \in \mathbb{R}^{B \times C \times h \times w}\) and radar features \(r_L \in \mathbb{R}^{B \times C \times w}\) as input, fusing them column by column. Because camera and radar are horizontally aligned by calibration, we fuse only along the height direction and treat the one-dimensional radar feature as a depth prior for each image column. For each batch index $b$ and column $u$, we extract the image column $i_L[b,:,1\ldots h,u] \in \mathbb{R}^{C \times h}$, transpose it to an $h \times C$ sequence, and take the radar vector $r_L[b,:,u] \in \mathbb{R}^{C}$, which we replicate along the height axis $h$ times to obtain an $h \times C$ tensor. Concatenating the image and radar tensors along the channel dimension gives an $h \times 2C$ token sequence for that column, and stacking all columns produces $(B \cdot w, h, 2C)$ tokens. For each column, we apply a height-wise self-attention block to its $h \times 2C$ tokens: a linear projection from $2C$ to $C$, multi-head self-attention along the height dimension, and a feed-forward network. Because all columns share weights and are processed independently, the Height Fusion Block has time complexity \(\mathcal{O}(wh^2)\). The output has shape $(B \cdot w, h, C)$, which we reshape to $(B, C, h, w)$ and pass through a lightweight convolutional module to restore local two-dimensional context. Height-wise attention learns to place the radar prior at consistent vertical positions in each column and to preserve plausible depth values around those positions.

\begin{figure}[t]
 \centering
 \includegraphics[width=0.99\columnwidth,trim=0 0 0 0,clip]{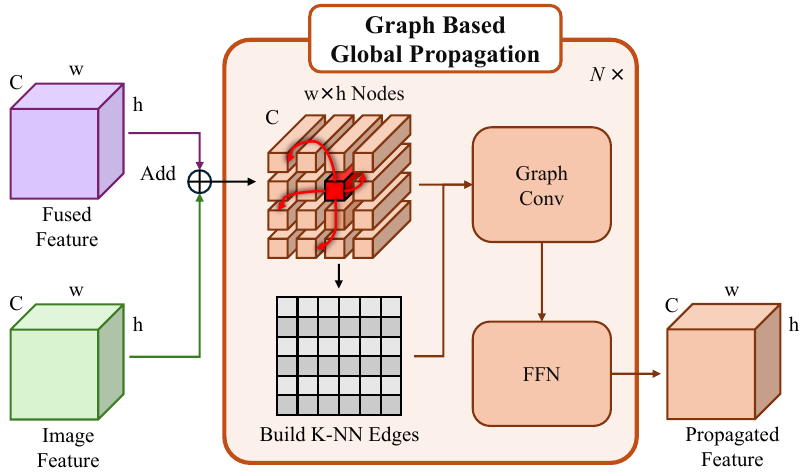}
 \vspace{-20pt}
\caption{\textbf{Graph-based depth propagation.} We sum the fused features and final image features, build a sparse $K$-NN graph over spatial tokens, apply MRConv followed by an FFN to propagate cues, and output a $C\times h\times w$ feature map.}
 \label{fig:gbp}
\end{figure}

\subsection{Graph‐Based Depth Propagation}
We propagate global depth context using VIG layers~\cite{han2022vig} equipped with Max-Relative Graph Convolution (MRConv)~\cite{li2019deepgcns}; see Fig.~\ref{fig:gbp}. We sum the fused and image features, \(X_0 = X_{\mathrm{fused}} + i_L \in \mathbb{R}^{C \times h \times w}\), and reshape them into tokens \(\{x_p\}_{p=1}^{M}\) with \(M = h \cdot w\) and \(x_p \in \mathbb{R}^{C}\). Each token \(x_p\) corresponds to one spatial location in the latent feature map, so each graph node represents a single pixel.

For each image and layer \(\ell\), we build a $K$-NN graph in feature space. We L2-normalize features along the channel dimension and find the nearest neighbors by $\ell_2$ distance to obtain a (possibly dilated) $K$-NN graph. We keep self-loops for all nodes. Given the neighbor set \(\mathcal{N}^{(\ell)}_p\), MRConv forms a max-relative message
\begin{equation}
m_p^{(\ell)}=\max_{q\in \mathcal{N}^{(\ell)}_p}\bigl(x_q^{(\ell)}-x_p^{(\ell)}\bigr),
\end{equation}
and updates \(x_p^{(\ell)}\) by concatenating \([x_p^{(\ell)}\| m_p^{(\ell)}]\), applying a $1{\times}1$ projection $(2C\rightarrow C)$ and a lightweight feed-forward block, both with residual connections. Stacking $N$ layers yields \(X_N \in \mathbb{R}^{C \times h \times w}\).

Feature aggregation in each layer scales as $\mathcal{O}(MK)$, while the dense $K$-NN search costs $\mathcal{O}(M^2)$. An $N$-layer graph backbone therefore has complexity $\mathcal{O}(N(M^2 + MK))$.

Intuitively, each node exchanges depth cues with its $K$ most similar nodes in feature space, which often lie on the same physical object even if they are far apart in the image. Information first propagates within local neighborhoods and, after $N$ layers, can travel across the entire scene. This gives the GNN a large, data-adaptive receptive field that lets radar-supported and LiDAR-supervised pixels propagate information into unsupervised regions. The empirical latency-accuracy trade-off as $N$ increases is summarized in Table~\ref{tab:gnn_layers}.

\begin{table*}[t]
  \centering
  \caption{Comparison of radar-camera depth estimation methods on nuScenes}
  \label{tab:comparison}
  \begin{threeparttable}
    \setlength{\tabcolsep}{4pt}
    \begin{tabular*}{\textwidth}{@{\extracolsep{\fill}}
      l      
      c      
      c c c  
      S[table-format=3.1]                
      *{3}{S[table-format=4.1]}          
      *{3}{S[table-format=4.1]}          
      S[table-format=1.3]                
      S[table-format=1.3]                
      @{}}
      \toprule
      Model & GT & Radar & Images & Auxiliary & {Time  $\downarrow$}
            & \multicolumn{3}{c}{MAE (mm) $\downarrow$}
            & \multicolumn{3}{c}{RMSE (mm) $\downarrow$}
            & {AbsRel $\downarrow$}
            & {log10 $\downarrow$} \\
      \cmidrule(lr){7-9}\cmidrule(lr){10-12}
            &     & Frames &        & Annotations  & {(ms)}    
            & {0-50} & {0-70} & {0-80}
            & {0-50} & {0-70} & {0-80}
            & {0-70} & {0-70}
            \\
      \midrule
      RC-PDA~\cite{long2021radar}   & 25     & 5 & 3 & LS, 3B
        & 411.6
        & 2225.0 & 3326.1 & 3713.6
        & 4156.5 & 6700.6 & 7692.8
        & \multicolumn{1}{c}{--}
        & \multicolumn{1}{c}{--} \\
      DORN~\cite{DORN_radar}        & interp & 15 & 1 & \ding{55}
        & 247.9
        & 1926.6 & 2380.6 & 2467.7
        & 4124.8 & 5252.7 & 5554.3
        & \multicolumn{1}{c}{--}
        & \multicolumn{1}{c}{--} \\
      Singh \emph{et al.}~\cite{singh2023depth} & 161   & 1 & 1 & P
        & 283.4
        & 1727.7 & 2073.2 & 2179.3
        & 3746.8 & 4590.7 & 4898.7
        & \multicolumn{1}{c}{--}
        & \multicolumn{1}{c}{--} \\
      Li \emph{et al.}~\cite{li2024sparse} & 1      & 1 & 1 & S
        & \multicolumn{1}{c}{\underline{51.1}}
        & 1524.5 & 1822.9 & 1927.0
        & 3567.3 & 4303.6 & 4609.6
        & \multicolumn{1}{c}{--}
        & \multicolumn{1}{c}{--} \\
      CaFNet~\cite{sun2024cafnet}   & 161    & 1 & 1 & P
        & 115.9
        & 1674.0 & 2010.0 & 2109.0
        & 3674.0 & 4493.0 & 4765.0
        & \multicolumn{1}{c}{0.101}
        & \multicolumn{1}{c}{0.040} \\
      GET-UP~\cite{sun2025getup}    & 161    & 1 & 1 & P
        & \multicolumn{1}{c}{587.6}
        & \multicolumn{1}{c}{\textbf{1241.0}}
        & \multicolumn{1}{c}{\textbf{1541.0}}
        & \multicolumn{1}{c}{\textbf{1632.0}}
        & \multicolumn{1}{c}{\textbf{2857.0}}
        & \multicolumn{1}{c}{\textbf{3657.0}}
        & \multicolumn{1}{c}{\textbf{3932.0}}
        & \multicolumn{1}{c}{\underline{0.075}}
        & \multicolumn{1}{c}{\textbf{0.032}} \\
      LiRCDepth~\cite{sun2025lircdepth} & 161 & 1 & 1 & P
        & 60.3
        & 1514.0 & 1898.0 & 2009.0
        & 3330.0 & 4300.0 & 4617.0
        & \multicolumn{1}{c}{0.095}
        & \multicolumn{1}{c}{0.036} \\
      \midrule
      \textbf{JustDepth (Ours)} & 1 & 1 & 1 & \ding{55}
        & \multicolumn{1}{c}{\textbf{14.8}}
        & \multicolumn{1}{c}{\underline{1324.2}}
        & \multicolumn{1}{c}{\underline{1674.0}}
        & \multicolumn{1}{c}{\underline{1771.3}}
        & \multicolumn{1}{c}{\underline{3285.5}}
        & \multicolumn{1}{c}{\underline{4292.9}}
        & \multicolumn{1}{c}{\underline{4602.0}}
        & \multicolumn{1}{c}{\textbf{0.074}}  
        & \multicolumn{1}{c}{\underline{0.033}}  
        \\
      \bottomrule
    \end{tabular*}
    \begin{tablenotes}[flushleft]
      \scriptsize
      \item Experiments use images of resolution $900\times1600$ with 100 radar points. Inference time is measured on an NVIDIA RTX~4070~Ti. Evaluations on the nuScenes~\cite{caesar2020nuscenes} test set are reported at 50\,m, 70\,m, and 80\,m. We compare against existing methods~\cite{long2021radar,DORN_radar,singh2023depth,li2024sparse,sun2024cafnet,sun2025getup,sun2025lircdepth}. 
\textbf{Abbreviations:} LS = LiDAR Semantic Segmentation, 3B = 3D Bounding Box, P = Image Panoptic Segmentation, S = Image Semantic Segmentation, \ding{55} = no auxiliary annotations.
    \end{tablenotes}
  \end{threeparttable}
\end{table*}

 \vspace{-0.2cm}

\subsection{Depth Decoder}
The depth decoder follows a U-Net-style upsampling strategy~\cite{ronneberger2015u} that fuses globally aggregated features from the graph backbone with intermediate image features. Using the final GNN output together with image encoder skip features \(\{i_0, i_1, i_2, i_3\}\), we perform four upsampling stages. At each stage, we upsample the current feature map by a factor of two, concatenate it with the corresponding skip feature, and pass the result through a lightweight convolutional block to blend features. After these stages, a final upsampling and a final convolution produce the depth prediction
\(
D_{\mathrm{pred}}\;\in\;\mathbb{R}^{B\times1\times H\times W}
\)
at the original image resolution.
During training,   \(D_{\mathrm{pred}}\)  is supervised against the single-scan LiDAR ground truth \(D_{\mathrm{gt}}\) by combining two complementary losses:

\noindent\textbf{Depth Regression Loss} 
We utilize a masked Smooth \(L_1\) (Huber) loss over valid pixels \((x,y)\in\Omega\):
\begin{equation}
    \mathcal{L}_{\mathrm{depth}}
    = \frac{1}{|\Omega|}\sum_{(x,y)\in\Omega}
       \mathrm{smooth}_\beta\bigl(D_{\mathrm{pred}}(x,y)-D_{\mathrm{gt}}(x,y)\bigr).
\end{equation}

\noindent\textbf{Edge-Aware Smoothness Loss}
We penalize first-order depth variation using weights derived from image gradients, assigning smaller penalties near strong intensity transitions.
Let the forward differences be
$\nabla_x D_{\mathrm{pred}}(x,y)=D_{\mathrm{pred}}(x{+}1,y)-D_{\mathrm{pred}}(x,y)$ and
$\nabla_y D_{\mathrm{pred}}(x,y)=D_{\mathrm{pred}}(x,y{+}1)-D_{\mathrm{pred}}(x,y)$.
From a grayscale image we compute a Sobel gradient magnitude $S(x,y)$ and set
$\omega(x,y)=\exp(-S(x,y))$.
The loss encourages depth to change smoothly in low-texture areas, but still allows abrupt depth changes along image edges. 
\vspace{-10pt}
\begin{equation}
    \begin{aligned}
    &\mathcal{L}_{\mathrm{smooth}} = \\
    &\frac{1}{|\Omega|}\!\sum_{(x,y)\in\Omega}
    \Big(|\nabla_x D_{\mathrm{pred}}(x,y)|+|\nabla_y D_{\mathrm{pred}}(x,y)|\Big)\,\omega(x,y).
    \end{aligned}
\vspace{-16pt}
\end{equation}

\begin{figure}[t]
  \centering
  \subfloat[]{%
    \includegraphics[width=0.32\linewidth]{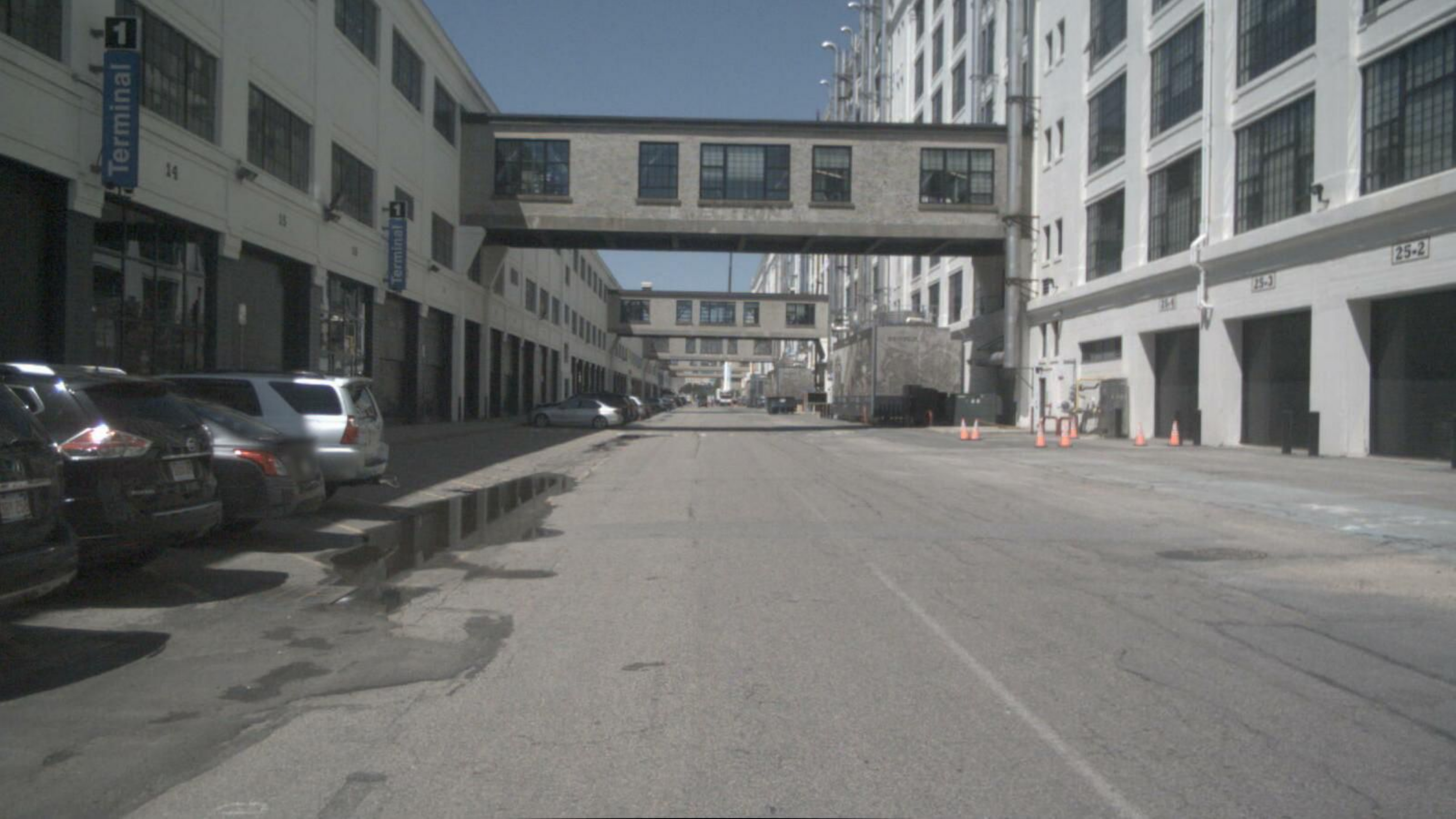}%
    \label{fig:aug_a}}
  \hfil
  \subfloat[]{%
    \includegraphics[width=0.32\linewidth]{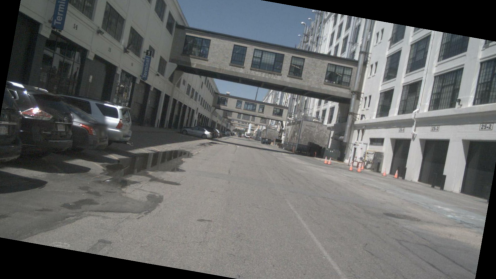}%
    \label{fig:aug_b}}
  \hfil
  \subfloat[]{%
    \includegraphics[width=0.32\linewidth]{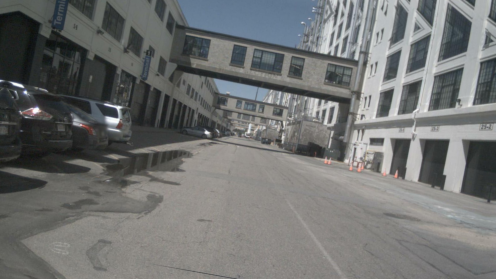}%
    \label{fig:aug_c}}
  \hfil
  \subfloat[]{%
    \includegraphics[width=0.32\linewidth]{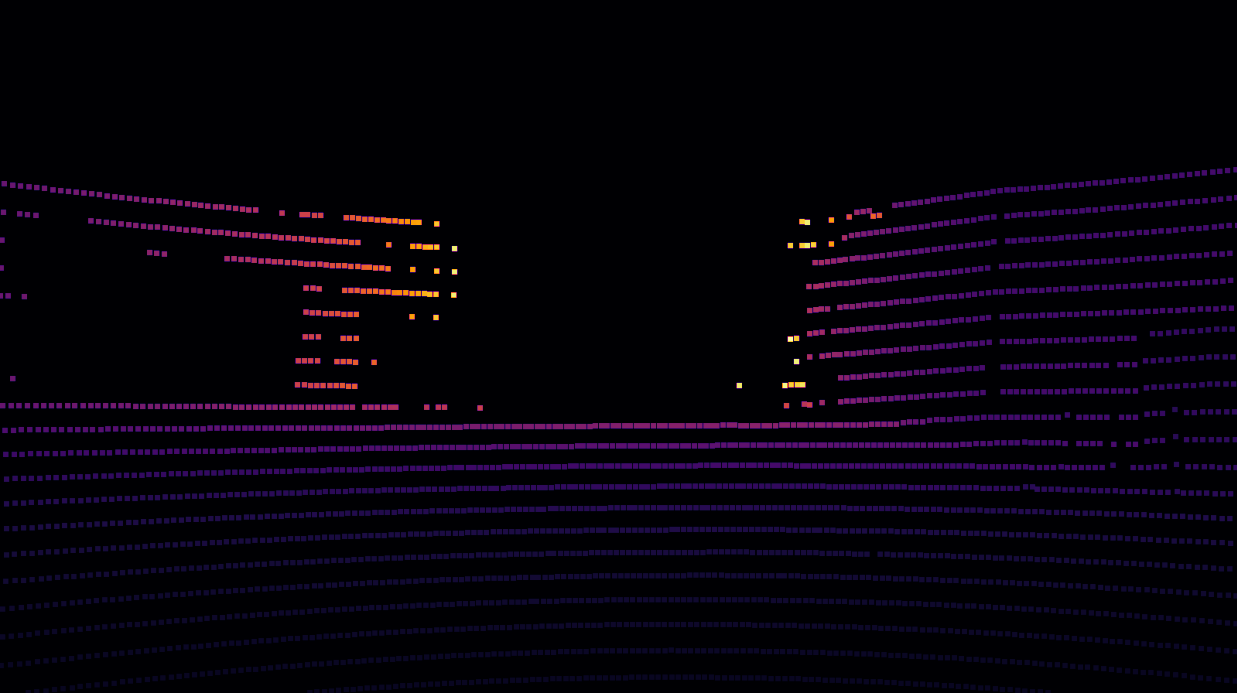}%
    \label{fig:aug_d}}
    \hfil
  \subfloat[]{%
    \includegraphics[width=0.32\linewidth]{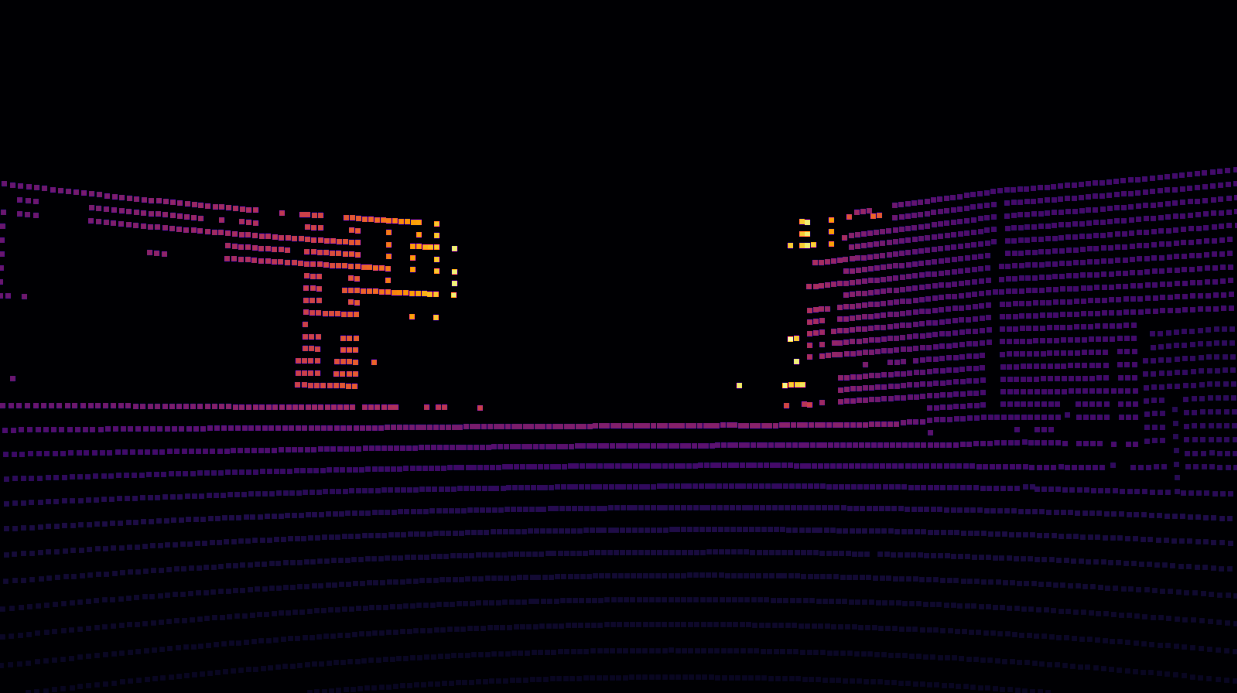}%
    \label{fig:aug_e}}
    \hfil
  \subfloat[]{%
    \includegraphics[width=0.32\linewidth]{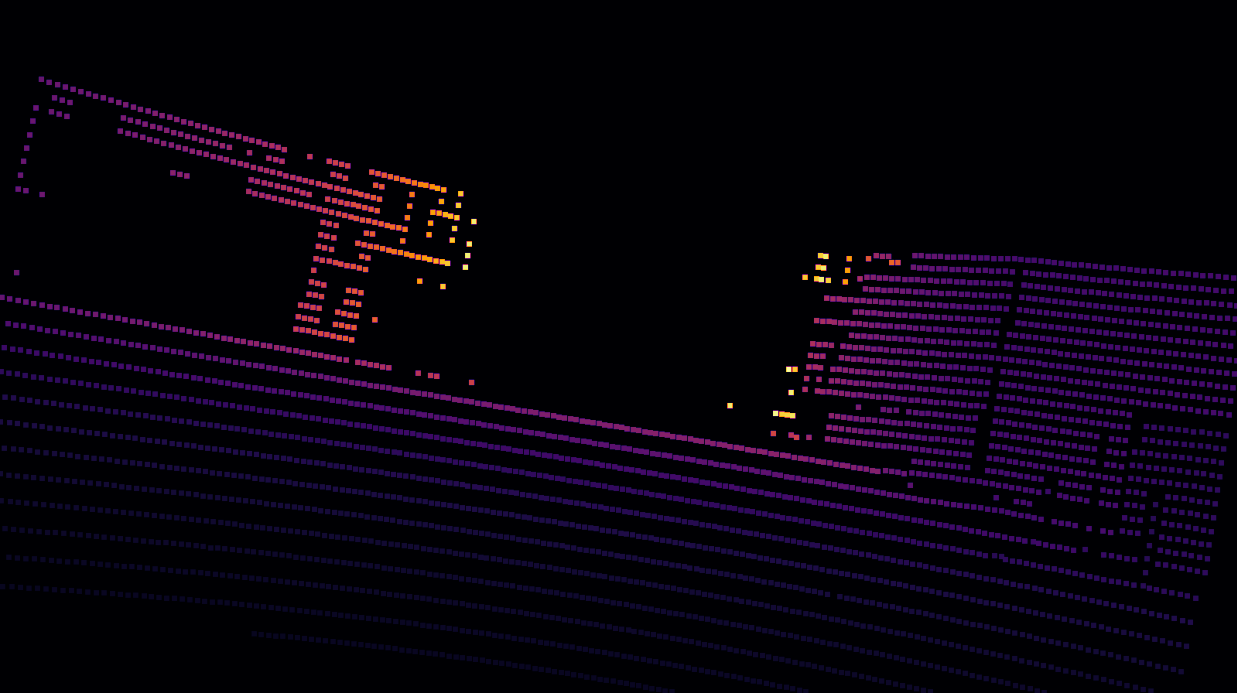}%
    \label{fig:aug_f}}
    \vspace{-5pt}
  \caption{\textbf{Rotation augmentation with reflection padding.} (a) original RGB image, (b) rotation with zero padding, (c) rotation with reflection padding. (d) raw single-sweep LiDAR projection, (e) LiDAR after point upsampling, and (f) LiDAR after applying both upsampling and rotation.}
  \label{fig:rotation_aug}
\end{figure}

\subsection{Confidence Decoder (Training Only)}
During training, JustDepth predicts a radar confidence map from the fused feature
\(X_{\mathrm{fused}}\in\mathbb{R}^{B\times C\times h\times w}\) using a lightweight auxiliary head.
The confidence decoder takes \(X_{\mathrm{fused}}\) as input and produces a per-pixel confidence logit map.
It first applies a small number of convolutional blocks at resolution \((h\times w)\) to adapt the feature representation.
Then, it performs a sequence of bilinear upsampling stages that gradually increase the spatial resolution to \((H\times W)\).
A final convolution projects the feature to a single-channel logit map
\(Z\in\mathbb{R}^{B\times1\times H\times W}\),
where each logit \(Z_{b,1,y,x}\) encodes how strongly the model believes that the depth prediction at pixel \((x,y)\) is supported by radar.
This branch is used only as an auxiliary supervision signal and is discarded at inference, so it does not affect test-time latency.

\noindent\textbf{Ground‐truth Generation.}
We first interpolate the sparse LiDAR returns into a dense depth map
\(\tilde D_{\mathrm{gt}}\in\mathbb{R}^{H\times W}\) using grid-based interpolation.
We also build a visibility mask \(M\in\{0,1\}^{H\times W}\) around each projected LiDAR point. Pixels within a radius \(\rho\) are marked valid and all other pixels are ignored in the confidence loss.
For each radar return at image column \(u\) with measured depth \(d_r(u)\), a vertical strip is: 
\begin{equation}
\Omega(u)
=\bigl\{(x,y)\,\big|\,|x-u|\le \tfrac{w_r}{2},\;y=1,\dots,H\bigr\},
\end{equation}
where \(w_r\) is the strip width in pixels.
The confidence target at pixel \((x,y)\) in batch \(b\) is then
\begin{equation}
\begin{aligned}
\hat Z_{b,1,y,x}
= \max_{u\in\mathcal U}\;
&\mathbf{1}\!\left\{(x,y)\in\Omega(u)\right\}\;
\mathbf{1}\!\left\{M(x,y)=1\right\} \\
&\times\mathbf{1}\!\left\{\,\bigl|\tilde D_{\mathrm{gt}}(x,y)-d_r(u)\bigr|<\epsilon\right\},
\end{aligned}
\end{equation}
where \(\mathcal U\) indexes the radar columns and \(\mathbf{1}\{\cdot\}\) is the indicator function.
The product of the three indicator terms sets the target to one only when the pixel lies inside the strip, has a valid LiDAR depth, and the LiDAR and radar depths agree.
The outer maximum runs over all radar columns and sets the target to one when at least one column satisfies these conditions at \((x,y)\).
This produces a binary mask \(\hat Z\in\mathbb{R}^{B\times1\times H\times W}\) that marks pixels whose LiDAR depth is consistent with at least one nearby radar measurement.

\noindent\textbf{Loss.} 
Rather than applying a sigmoid to \(Z\), we directly optimize with a \emph{BCEWithLogits} loss. Denoting the logit at \((b,y,x)\) by \(Z_{b,1,y,x}\) and the target by \(\hat Z_{b,1,y,x}\), we minimize
\begin{equation}
\mathcal{L}_{\mathrm{conf}}
=\;\mathrm{BCEWithLogits}\bigl(Z,\;\hat Z\bigr),
\end{equation}
which penalizes the predicted logits against the binary confidence targets without requiring an explicit sigmoid activation.

\subsection{LiDAR Distribution Leakage Mitigation}
JustDepth is trained with single-scan LiDAR only. This avoids the misalignment and motion artifacts of multi-frame fusion~\cite{long2021radar,singh2023depth,li2024radarcam,sun2024cafnet,sun2025getup,sun2025lircdepth}, but sparse supervision can imprint the scanner’s sampling pattern, such as LiDAR Distribution Leakage (LDL), and produce stripe artifacts~\cite{li2024sparse}. To mitigate this while staying single-frame, we use point upsampling and rotation with reflection padding.

\noindent\textbf{Point Upsampling.}
We densify the supervision by adding midpoints between suitable pairs of projected LiDAR samples.
We build a KD-tree over 2D image coordinates and enumerate all point pairs whose Euclidean distance is at most $\rho$.
For two local neighbors $(x_i,y_i,d_i)$ and $(x_j,y_j,d_j)$, we retain pairs that are horizontally close
($\lvert x_i - x_j\rvert \le \tau_x$), vertically separated so they lie on different scanlines
($\lvert y_i - y_j\rvert \ge \tau_y$), and have similar depth ($\lvert d_i - d_j\rvert < \tau_d$)
For each such pair we insert an additional LiDAR sample at the midpoint in image space and assign it the average depth $(d_i + d_j)/2$.

\noindent\textbf{Rotation with Reflection Padding.}
We apply a rotation augmentation to the RGB image, LiDAR, and radar with the same random angle in $[-\Theta,\Theta]$.  For LiDAR and radar, we rotate the 2D projected point coordinates in the image plane around the image center. The rotated LiDAR points are then rasterized into a sparse depth map, and the rotated radar points are collapsed along the vertical direction to form the 1D scan. For the RGB image, we rotate the image itself and apply reflection padding. Rotation without padding produces black borders, as shown in Fig.~\ref{fig:rotation_aug}, that let the model infer the rotation angle and encourage shortcut learning. Such shortcuts cause the network to rely on border artifacts instead of learning to be robust to the LiDAR sampling pattern, which reduces the effectiveness of the rotation augmentation for mitigating LDL.

A single LiDAR sweep produces nearly horizontal scanlines with large gaps in the image plane. Crop-based augmentation, as in Li \emph{et al.}~\cite{li2024sparse}, moves these scanlines up and down but keeps their direction horizontal. Our rotation augmentation instead applies a different random angle to each training sample. The projected scanlines then appear with many directions across the dataset, so the model must produce depth that is consistent under changing stripe directions instead of reproducing a single fixed pattern. Point upsampling reduces the spacing between neighboring scanlines by inserting additional samples on object surfaces, which spreads supervision over more pixels and further weakens the incentive to maintain a stable stripe pattern tied to the original LiDAR channels.

\begin{figure*}[!t]
\centering
\vspace{5pt}
\begingroup
\footnotesize
\setlength{\tabcolsep}{0pt} 
\renewcommand{\arraystretch}{1.0}
\begin{tabular*}{0.99\textwidth}{
  @{\hspace{0.003\textwidth}}
  >{\centering\arraybackslash}m{0.16233\textwidth}
  @{\hspace{0.002\textwidth}}
  >{\centering\arraybackslash}m{0.16233\textwidth}
  @{\hspace{0.002\textwidth}}
  >{\centering\arraybackslash}m{0.16233\textwidth}
  @{\hspace{0.002\textwidth}}
  >{\centering\arraybackslash}m{0.16233\textwidth}
  @{\hspace{0.002\textwidth}}
  >{\centering\arraybackslash}m{0.16233\textwidth}
  @{\hspace{0.002\textwidth}}
  >{\centering\arraybackslash}m{0.16233\textwidth}
  @{\hspace{0.003\textwidth}}
}
\textbf{Image} &
\textbf{LiDAR GT} &
\textbf{Singh \emph{et al.}~\cite{singh2023depth}} &
\textbf{Li \emph{et al.}~\cite{li2024sparse}} &
\textbf{GET\text{-}UP~\cite{sun2025getup}} &
\textbf{Ours} \\
\end{tabular*}
\endgroup

\includegraphics[width=0.99\textwidth,trim=0 0 0 0,clip]{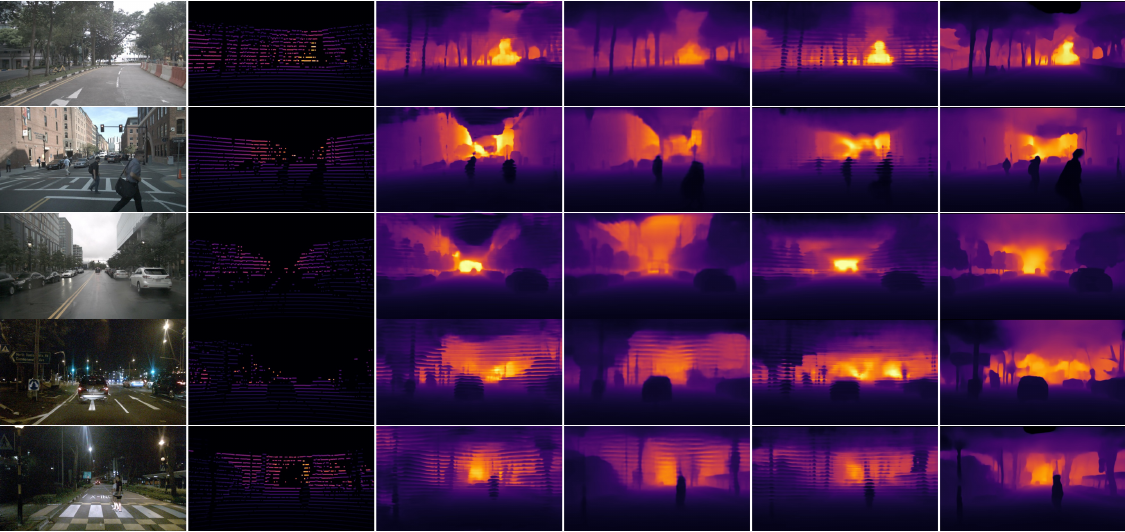}

\vspace{-10pt}
\caption{\textbf{Qualitative depth estimation comparison.} Each row shows the input RGB image, sparse single‐scan LiDAR ground truth, and predicted depth maps from Singh \emph{et al.}~\cite{singh2023depth}, Li \emph{et al.}~\cite{li2024sparse}, GET-UP~\cite{sun2025getup}, and the proposed method with an 8‐layer GNN. We include diverse scenarios from the nuScenes test set to demonstrate that JustDepth produces more complete and artifact‐free depth maps.}
\label{fig:qualitative}
\end{figure*}

\section{Experiments}

\subsection{Dataset and Evaluation Metrics}
We evaluate JustDepth on the nuScenes dataset~\cite{caesar2020nuscenes}, which comprises 1000 scenes collected in urban driving environments. Following common practice, we use 700 scenes for training, 150 scenes for validation, and 150 scenes for testing. Each sample includes synchronized RGB images, mmWave radar point clouds, and single‐scan LiDAR depth maps. Inputs are $900{\times}1600$ for both training and evaluation. Depth predictions are compared against LiDAR ground truth using the following error metrics: MAE, RMSE, AbsRel, and log10. MAE and RMSE are reported over the distance intervals 0-50\,m, 0-70\,m, and 0-80\,m, while AbsRel and log10 are reported over 0-70\,m.

\subsection{Implementation Details}
The model is trained for 200 epochs on two NVIDIA RTX~3090 GPUs with a per-GPU batch size of 8 (16 total). We use AdamW~\cite{loshchilov2019decoupled} optimizer with a cosine-annealed learning rate schedule~\cite{loshchilov2017sgdr}. Data augmentation includes random horizontal flips, color jitter, point upsampling, and synchronized rotations of the image, radar, and LiDAR with reflection padding on the image only. For point upsampling, we use a KD-tree radius $\rho{=}48$ pixels and keep pairs that satisfy $\tau_x{=}4$ pixels, $\tau_y{=}16$ pixels, and $\tau_d{=}0.2$\,m. Rotation angles are sampled uniformly between $[-10^\circ,10^\circ]$. The GNN has 8 layers with $K{=}9$ neighbors per node and a dilated $K$-NN schedule across depth. Inference time is measured on an NVIDIA RTX~4070~Ti. The training objective is a weighted sum of depth regression, edge-aware smoothness, and confidence terms, $\mathcal{L}_{\mathrm{total}}=\mathcal{L}_{\mathrm{depth}}+\lambda_{\mathrm{smooth}}\,\mathcal{L}_{\mathrm{smooth}}+\lambda_{\mathrm{conf}}\,\mathcal{L}_{\mathrm{conf}}$, where we set $\lambda_{\mathrm{smooth}}{=}0.1$ and $\lambda_{\mathrm{conf}}{=}10.0$.

\subsection{Quantitative Results}
Table~\ref{tab:comparison} compares latency and accuracy across competing methods on nuScenes~\cite{caesar2020nuscenes}. JustDepth is much more efficient while keeping accuracy competitive. It processes a frame in 14.8\,ms, about $39.7\times$ faster than GET-UP~\cite{sun2025getup}, which is important for on-board deployment. The accuracy gap is modest: JustDepth has only 8.54\% higher MAE than GET-UP~\cite{sun2025getup}, while the architecture is significantly simpler. In terms of relative error, JustDepth achieves the best AbsRel (0.074 vs. 0.075 for GET-UP), which means smaller depth errors with respect to the true distance, especially for nearby objects. GET-UP~\cite{sun2025getup} merges 161 LiDAR frames for dense supervision and uses panoptic segmentation to filter dynamic objects, whereas our model uses a single LiDAR sweep with no auxiliary annotations. This reduces system complexity and data requirements without sacrificing performance.

JustDepth also improves over the previously fastest method by Li \emph{et al.}~\cite{li2024sparse} in both speed and accuracy. Their model uses sparse single-scan LiDAR with semantic segmentation masks and runs at about $51$\,ms per frame. Our model is more than $3\times$ faster at 14.8\,ms and also achieves better depth accuracy. At the full 0-80\,m range, JustDepth reduces MAE by about 8.08\% compared to Li \emph{et al.}~\cite{li2024sparse}, even though it does not use segmentation guidance. These results show that JustDepth delivers high efficiency and strong accuracy with weaker data dependence, which is desirable for time-critical systems.

\subsection{Qualitative Results}
Fig.~\ref{fig:qualitative} shows depth predictions on nuScenes~\cite{caesar2020nuscenes} scenes ranging from day-time to dark night-time environments. Compared with other methods, JustDepth generates smoothly varying depth without the horizontal stripe artifacts common in sparse LiDAR supervision. For example, Li \emph{et al.}~\cite{li2024sparse} supervise on a single LiDAR sweep and try to mitigate LiDAR Distribution Leakage using augmentations and image semantic supervision, but their results still exhibit visible stripe artifacts under low-light conditions. In contrast, JustDepth maintains artifact-free depth maps even at night. Scene structures and boundaries remain sharp and accurate, showing that our fusion of sparse radar returns with image features produces high-fidelity, stripe-free reconstructions across all lighting conditions.

\begin{table}[t]
  \centering
  \caption{Comparison of LDL metrics}
  \label{tab:stripe_metrics}
  \begin{threeparttable}
    \begin{tabular*}{\linewidth}{@{\extracolsep{\fill}} lccc @{}}
      \toprule
      Method & OMAE (mm) $\downarrow$ & VGM $\downarrow$ & VHGR $\downarrow$ \\
      \midrule
      Singh \emph{et al.}~\cite{singh2023depth} & 2105.3  & 0.2145  & 0.3813 \\
      Li \emph{et al.}~\cite{li2024sparse}      & \textbf{1881.0} & 0.1146  & 0.1868 \\
      GET-UP~\cite{sun2025getup}               & 2209.0  & 0.2565  & 0.5210 \\
      JustDepth (Ours)                         & 2119.9  & \textbf{0.1090} & \textbf{0.1767} \\
      \bottomrule
    \end{tabular*}
    \begin{tablenotes}[flushleft]
      \scriptsize
  \item OMAE follows Li \emph{et al.}~\cite{li2024sparse}, while VGM and VHGR quantify LDL-related striping artifacts. All metrics are reported on the nuScenes~\cite{caesar2020nuscenes} test set.
    \end{tablenotes}
  \end{threeparttable}
\end{table}

\subsection{Gradient Metrics}
Li \emph{et al.}~\cite{li2024sparse} introduced the object-level MAE (OMAE) as a measure of LiDAR Distribution Leakage, but it is restricted to LiDAR samples within the selected semantic masks. To capture scanline artifacts over the whole image, we use a gradient-based metric comparing vertical and horizontal depth changes.
Depth is clipped to $0$-$50$\,m for all methods, and we crop a central band of the image to rows $y\!\in\![200,699]$ for $900{\times}1600$ inputs.
For each depth map $D \in \mathbb{R}^{H \times W}$, 
\vspace{-5pt}
\begin{equation}
\begin{aligned}
G_v(D)&=\frac{1}{(H-1)W}\!\sum_{y=1}^{H-1}\sum_{x=1}^{W}\bigl|D_{y+1,x}-D_{y,x}\bigr|,\\
G_h(D)&=\frac{1}{H(W-1)}\!\sum_{y=1}^{H}\sum_{x=1}^{W-1}\bigl|D_{y,x+1}-D_{y,x}\bigr|.
\end{aligned}
\vspace{-5pt}
\end{equation}
$G_v(D)$ and $G_h(D)$ denote the mean absolute vertical and horizontal depth differences in the depth map, and we define the Vertical Gradient Magnitude as $\mathrm{VGM}(D)=G_v(D)$.

\begin{table}[t]
  \centering
  \caption{Ablation of GNN depth and convolutional baseline}
  \label{tab:gnn_layers}
  \begin{threeparttable}
    \begin{tabular*}{\linewidth}{@{\extracolsep{\fill}} lcccc @{}}
      \toprule
      & \multicolumn{3}{c}{\textbf{GNN Layers}} & \textbf{CNN Layers} \\
      \cmidrule(lr){2-4} \cmidrule(lr){5-5}
      \textbf{Metric} & \textbf{0} & \textbf{8} & \textbf{16} & \textbf{8} \\
      \midrule
      FLOPs (G) $\downarrow$
        & \textbf{33.67} & 42.58 & 51.49 & 43.29 \\
      Params (M) $\downarrow$
        & \textbf{9.46} & 15.81 & 22.15 & 16.32 \\
      MAE$_{70}$ (mm) $\downarrow$
        & 1774.1 & 1674.0 & \textbf{1644.6} & 1768.8 \\
      RMSE$_{70}$ (mm) $\downarrow$
        & 4396.3 & 4292.9 & \textbf{4259.7} & 4417.3 \\
      \bottomrule
    \end{tabular*}
    \begin{tablenotes}[flushleft]
      \scriptsize
\item The number of GNN layers ($N{=}0,8,16$) varies against
      an 8-layer convolutional baseline. FLOPs and parameter counts are measured for JustDepth without the confidence decoder. Results are reported on the nuScenes~\cite{caesar2020nuscenes} test set (0-70\,m).
    \end{tablenotes}
  \end{threeparttable}
\end{table}

\begin{table}[t]
  \centering
  \caption{Ablation of key components of JustDepth}
  \label{tab:ablation_module}
  \begin{threeparttable}
    \begin{tabular*}{\linewidth}{@{\extracolsep{\fill}} c c c c c @{}}
      \toprule
      \multicolumn{3}{c}{\textbf{Components}} & \multicolumn{2}{c}{\textbf{Error 0-70\,m} $\downarrow$} \\
      \cmidrule(lr){1-3} \cmidrule(lr){4-5}
      \textbf{Conf Dec} & \textbf{Self-Attn} & \textbf{GNN} & \textbf{MAE$_{70}$} & \textbf{RMSE$_{70}$} \\
      \midrule
      \ding{51} & \ding{51} & \ding{51} & \textbf{1674.0} & \textbf{4292.9} \\
      \ding{55} & \ding{51} & \ding{51} & 1712.4 & 4377.9 \\
      \ding{51} & \ding{55} & \ding{51} & 1738.7 & 4414.7 \\
      \ding{51} & \ding{51} & \ding{55} & 1774.1 & 4396.3 \\
      \bottomrule
    \end{tabular*}
    \begin{tablenotes}[flushleft]
      \scriptsize
      \item Removing any component degrades performance. All settings are evaluated on the nuScenes~\cite{caesar2020nuscenes} test set (0-70\,m). 
    \end{tablenotes}
  \end{threeparttable}
\end{table}
Vertical-Horizontal Gradient Ratio (VHGR) is defined to capture the imbalance between vertical and horizontal depth changes:
\vspace{-12pt}
\begin{equation}
\mathrm{VHGR}(D)=\frac{G_v(D)-G_h(D)}{G_v(D)+G_h(D)+\varepsilon}.
\end{equation}
This ratio is near zero when vertical and horizontal gradients have similar magnitude, and becomes positive when vertical gradients dominate.
LiDAR Distribution Leakage produces horizontally aligned stripes, which increase vertical depth changes much more than horizontal ones and therefore lead to a higher VHGR.
On nuScenes~\cite{caesar2020nuscenes}, JustDepth attains the lowest VHGR among the compared methods (Table~\ref{tab:stripe_metrics}), while Li \emph{et al.}~\cite{li2024sparse} obtain the lowest OMAE.
However, OMAE is computed only at LiDAR samples inside selected semantic masks and averages depth within each mask weighted by the number of LiDAR points, so classes with many returns dominate the score and striping in sparse or non-selected regions is weakly reflected.
In our experiments, we therefore use OMAE as a complementary indicator of object-level depth bias, whereas VHGR is more suitable for characterizing LDL-related scanline artifacts over the entire depth map.

\subsection{Ablation Study}

\begin{table}[t]
  \centering
  \caption{Ablation of augmentations and smoothness}
  \label{tab:ablation_aug}
  \begin{threeparttable}
    \begin{tabular*}{\linewidth}{@{\extracolsep{\fill}} c c c c c c @{}}
      \toprule
      \multicolumn{3}{c}{\textbf{Components}} &
      \multicolumn{2}{c}{\textbf{Error 0-70\,m} $\downarrow$} &
      \multicolumn{1}{c}{\textbf{Gradient metric}} \\
      \cmidrule(lr){1-3} \cmidrule(lr){4-5} \cmidrule(lr){6-6}
      R & U & S & MAE & RMSE & VHGR \\
      \midrule
      \cmark & \cmark & \cmark & 1674.0   & 4292.9 & 0.1767 \\
      \xmark & \cmark & \cmark & \textbf{794.1} & \textbf{2701.6} & 0.8222 \\
      \cmark & \xmark & \cmark & 1662.1 & 4270.0 & 0.2113 \\
      \cmark & \cmark & \xmark & 1676.3 & 4311.1 & \textbf{0.1698} \\
      \bottomrule
    \end{tabular*}
    \begin{tablenotes}[flushleft]
      \scriptsize
      \item Ablation of rotation, point upsampling, and edge-aware smoothness on the nuScenes~\cite{caesar2020nuscenes} test set (0-70\,m). We report MAE, RMSE, and VHGR for each setting.
      (\textbf{R}: rotation with reflection padding,\quad
            \textbf{U}: point upsampling,\quad
            \textbf{S}: edge-aware smoothness loss.)
    \end{tablenotes}
  \end{threeparttable}
\end{table}

\begin{figure}[t]
\vspace{-0.5mm}
  \centering
  \subfloat[w/o rotation]{%
    \includegraphics[width=0.49\linewidth]{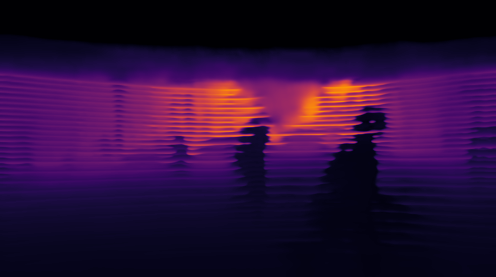}}
  \hfill
  \subfloat[w/o point upsampling]{%
    \includegraphics[width=0.49\linewidth]{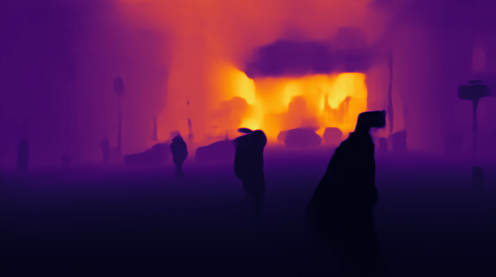}}
  \vspace{-6pt}
  \subfloat[w/o edge-aware smoothness loss]{%
    \includegraphics[width=0.49\linewidth]{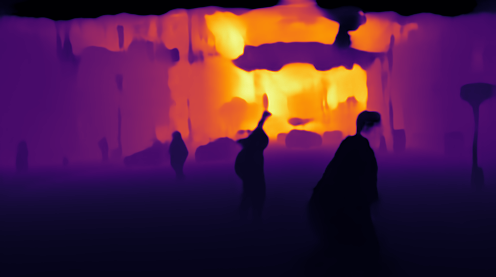}}
  \hfill
  \subfloat[full model]{%
    \includegraphics[width=0.49\linewidth]{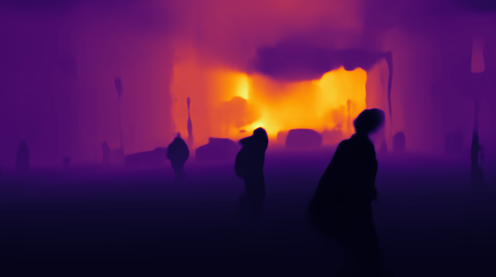}}
  \vspace{-2pt}
\caption{\textbf{Qualitative ablation of augmentations and smoothness.}
(a) without rotation with reflection padding,
(b) without point upsampling,
(c) without edge-aware smoothness loss,
and (d) with rotation, point upsampling, and edge-aware smoothness loss enabled, evaluated on the nuScenes test set.}
  \label{fig:aug_ablation_viz}
\end{figure}

\begin{table}[t]
  \centering
  \caption{Effect of rotation range on depth accuracy and striping}
  \label{tab:ablation_rot}
  \begin{threeparttable}
    \begin{tabular*}{\linewidth}{@{\extracolsep{\fill}} c c c c @{}}
      \toprule
      \multicolumn{1}{c}{\textbf{Rotation}} &
      \multicolumn{2}{c}{\textbf{Error 0-70\,m} $\downarrow$} &
      \multicolumn{1}{c}{\textbf{Gradient metric} $\downarrow$} \\
      \cmidrule(lr){1-1} \cmidrule(lr){2-3} \cmidrule(lr){4-4}
      $\Theta$ & MAE & RMSE & VHGR \\
      \midrule
      0   & \textbf{794.1}   & \textbf{2701.6} & 0.8222 \\
      5   & 1648.3           & 4278.5          & 0.2227 \\
      10  & 1674.0           & 4292.9          & 0.1767 \\
      15  & 1682.4           & 4306.2          & \textbf{0.1585} \\
      \bottomrule
    \end{tabular*}
    \begin{tablenotes}[flushleft]
      \scriptsize
    \item Ablation of the rotation range $\Theta$ on the nuScenes test set (0-70\,m); point upsampling and edge-aware smoothness are enabled in all settings.
    \end{tablenotes}
  \end{threeparttable}
\end{table}

\begin{table}[t]
  \centering
  \caption{Effect of different LDL mitigation strategies within JustDepth}
  \label{tab:ldl_aug}
  \begin{threeparttable}
    \begin{tabular*}{\linewidth}{@{\extracolsep{\fill}} lcccc @{}}
      \toprule
      Augmentation & MAE$_{70}$ $\downarrow$ & RMSE$_{70}$ $\downarrow$ & VGM $\downarrow$ & VHGR $\downarrow$ \\
      \midrule
      Li \emph{et al.}~\cite{li2024sparse} aug. 
        & \textbf{1586.6} & \textbf{4112.2} & 0.1688 & 0.2699 \\
      Ours (rot.~+ upsample) 
        & 1674.0 & 4292.9 & \textbf{0.1090} & \textbf{0.1767} \\
      \bottomrule
    \end{tabular*}
    \begin{tablenotes}[flushleft]
      \scriptsize
\item We compare Li \emph{et al.}~\cite{li2024sparse} augmentations with our rotation+upsampling under the same JustDepth setup, evaluated on the nuScenes test set (0-70\,m).
    \end{tablenotes}
  \end{threeparttable}
\end{table}

\noindent\textbf{Effect of GNN.}
Table~\ref{tab:gnn_layers} summarizes the trade-off between GNN depth, computational cost, and accuracy, and compares our graph backbone with a convolutional baseline of similar complexity.
As the number of GNN layers $N$ increases, both FLOPs and parameters grow roughly linearly while the depth error consistently decreases, showing that deeper graph propagation effectively improves global consistency. At similar FLOPs and parameter counts, the GNN backbone outperforms the 8-layer CNN baseline, so we adopt $N{=}8$ as a balanced operating point that offers clear gains over pure convolution while keeping the overall model compact.

\noindent\textbf{Key components.}
Table~\ref{tab:ablation_module} shows that removing any single component consistently degrades performance.
Omitting the GNN leads to the largest drop, highlighting the importance of non-local propagation.
Disabling self-attention in the Height Fusion Block also harms accuracy, indicating that column-wise contextual reasoning is beneficial.
The training-only Confidence Decoder yields a smaller but consistent improvement, and since it is discarded at inference, these gains come with zero test-time overhead.

\noindent\textbf{Augmentations \& smoothness.}
Table~\ref{tab:ablation_aug}, Fig.~\ref{fig:aug_ablation_viz}, and Table~\ref{tab:ablation_rot} reveal a structured trade-off between standard depth errors and LiDAR Distribution Leakage under single-sweep LiDAR supervision. Because MAE and RMSE are computed against sparse, scanline-structured LiDAR, the configuration without rotation ($\Theta{=}0^\circ$) can reduce these metrics by reproducing the LiDAR channel pattern, which yields the lowest MAE/RMSE but also the highest VHGR and strong horizontal striping in Fig.~\ref{fig:aug_ablation_viz}(a). When we enable rotation and point upsampling, the LiDAR scanlines are continuously perturbed and densified, discouraging overfitting to a fixed sampling pattern. The network is then driven toward smoother, more physically plausible depth that is consistent across stripe orientations, which increases MAE/RMSE but dramatically lowers VHGR and produces much cleaner depth maps. Importantly, since single-sweep LiDAR is itself sparse and noisy, even an oracle model that perfectly matches the true continuous scene depth cannot achieve zero MAE/RMSE against this GT, so a modest degradation in MAE/RMSE is an expected side effect of stronger LDL suppression rather than a contradiction. Overall, these ablations show that our LDL strategy deliberately trades a small amount of point-wise error for a substantial reduction in scanline artifacts, a behavior that is captured by VHGR and the qualitative results but not by MAE/RMSE alone.

Table~\ref{tab:ldl_aug} further clarifies the effect of the LDL mitigation strategy within JustDepth. When we replace our rotation and point upsampling with the crop-based Camera Intrinsics Disruption of Li \emph{et al.}~\cite{li2024sparse}, which combines random upsampling and random cropping of the image-radar-LiDAR pair, MAE and RMSE become slightly lower. However, VGM and VHGR increase by roughly 35\%, revealing markedly stronger vertical striping. This trade-off suggests that their disruption scheme mainly improves standard error metrics while tolerating more pronounced scanline artifacts, whereas our LDL strategy produces much cleaner depth maps at comparable error levels and is therefore better suited for practical deployment.

\section{Conclusion}
JustDepth is a single-stage radar-camera depth estimator trained solely on single-scan LiDAR supervision, without auxiliary annotations. A fixed-width radar encoding ensures constant-latency inference. A Height Fusion Block and a lightweight GNN propagate depth cues globally, while a training-only confidence head regularizes optimization with zero test-time cost. Although it is trained with sparse single-scan supervision, our use of augmentations substantially mitigates LDL, producing stripe-free depth and the lowest VHGR among compared methods.  JustDepth achieves real-time performance while maintaining competitive accuracy.

{
    \small
    \bibliographystyle{IEEEtran}
    \bibliography{main}
}

\end{document}